\documentclass[10pt,twocolumn,letterpaper]{article}

\usepackage{iccv}
\usepackage{times}
\usepackage{epsfig}
\usepackage{graphicx}
\usepackage{amsmath}
\usepackage{amssymb}
\usepackage{makecell}
\usepackage{multirow}
\usepackage{caption} 
\usepackage[breaklinks=true,bookmarks=false]{hyperref}

\iccvfinalcopy % *** Uncomment this line for the final submission

\def\iccvPaperID{****} % *** Enter the ICCV Paper ID here

\ificcvfinal\fi

\begin{document}

%%%%%%%%% TITLE
\title{TICaM: A Time-of-flight In-car Cabin Monitoring Dataset}

% \author{
%     \IEEEauthorblockN{Jigyasa Singh Katrolia, Bruno Mirbach, Ahmed El-Sherif, Hartmut Feld,\\Jason Rambach and Didier Stricker}\\
%     \IEEEauthorblockA{\normalsize Augmented Vision Department\\
%     \normalsize German Research for Artificial Intelligence\\
%     \normalsize Kaiserslautern, Germany\\
%     \{jigyasa\_singh.katrolia, bruno.mirbach, ahmed.elsherif, jason.rambach, didier.stricker\}@dfki.de}
%     }
%\author{Jigyasa Singh Katrolia\thanks{jigyasa\_singh.katrolia@dfki.de} %
%\and Bruno Mirbach\thanks{bruno.mirbach@dfki.de} %
%\and Ahmed El-Sherif\thanks{ahmed.elsherif@dfki.de}
%\and Hartmut Feld
%\and Jason Rambach\thanks{jason\_raphael.rambach@dfki.de}
%\and Didier Stricker\thanks{didier.stricker@dfki.de}\\\\
%\affiliation{German Research for Artificial Intelligence}}

\author{Jigyasa Singh Katrolia $\;\;\;\;\;$ Bruno Mirbach $\;\;\;\;\;$ Ahmed El-Sherif $\;\;\;\;\;$ Hartmut Feld \vspace{2pt}\\ $\;\;\;\;\;$ Jason Rambach $\;\;\;\;\;$ Didier Stricker \vspace{7pt} \\
DFKI - German Research Center for Artificial Intelligence\\
{\tt\small firstname.lastname@dfki.de}
}
% For a paper whose authors are all at the same institution,
% omit the following lines up until the closing ``}''.
% Additional authors and addresses can be added with ``\and'',
% just like the second author.
% To save space, use either the email address or home page, not both
% \and
% Second Author\\
% Institution2\\
% First line of institution2 address\\
% {\tt\small secondauthor@i2.org}

\maketitle
\ificcvfinal\thispagestyle{empty}\fi

%%%%%%%%% ABSTRACT
\begin{abstract}
   We present TICaM, a Time-of-flight In-car Cabin Monitoring dataset for  vehicle interior monitoring using a single wide-angle depth camera. Our dataset addresses the deficiencies of currently available in-car cabin datasets in terms of the ambit of labeled classes, recorded scenarios and provided annotations; all at the same time. We record an exhaustive list of actions performed while driving and provide for them multi-modal labeled images (depth, RGB and IR), with complete annotations for 2D and 3D object detection, instance and semantic segmentation as well as activity annotations for RGB frames. Additional to real recordings, we provide a synthetic dataset of in-car cabin images with same multi-modality of images and annotations, providing a unique and extremely beneficial combination of synthetic and real data for effectively training cabin monitoring systems and evaluating domain adaptation approaches. The dataset is available at https://vizta-tof.kl.dfki.de/.
\end{abstract}

%%%%%%%%% BODY TEXT
\section{Introduction}
\label{Intro}
With the advent of autonomous and driver-less vehicles, it is imperative to monitor the entire in-car cabin scene in order to realize active and passive safety functions, as well as comfort functions and advanced human-vehicle interfaces. Such car cabin monitoring systems typically involve a camera fitted in the overhead module of a car and a suite of algorithms to monitor the environment within a vehicle \cite{valeo,smarteye}. Ever growing performance of deep learning based computer vision methods has made it possible to monitor dynamic scenarios inside a car with high accuracy. To aid these monitoring systems, several in-car datasets exist to train deep leaning methods for solving problems like driver distraction monitoring, occupant detection or activity recognition \cite{Brain4Cars,HEH,abouelnaga2018DDR,drive_and_act_2019_iccv}. In the same vein, we present TICaM, an in-car cabin dataset of ~6.7K real time-of-flight depth images and 3.3k synthetic images with ground truth annotations for 2D and 3D object detection, and semantic and instance segmentation. In addition, we provide RGB video stream of driver and passenger activity along with activity annotations, totalling 123K frames. Our intention is to provide a comprehensive in-car cabin depth image dataset that addresses the deficiencies of currently available such datasets in terms of the ambit of labeled classes, recorded scenarios and provided annotations; all at the same time. 
%It can be used for developing machine learning based systems for extensive car cabin monitoring systems. 
\begin{figure}[h]
  	\centering
	\setlength{\tabcolsep}{2pt}
	\begin{tabular}{ccc}
		\includegraphics[width = 0.15\textwidth]{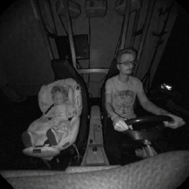} &
		\includegraphics[width = 0.15\textwidth]{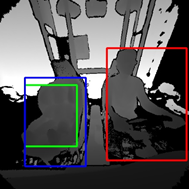} &
		\includegraphics[width = 0.15\textwidth]{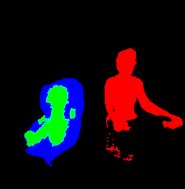} \\
		\includegraphics[width = 0.15\textwidth]{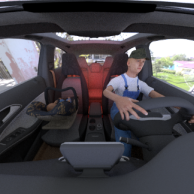} &
		\includegraphics[width = 0.15\textwidth]{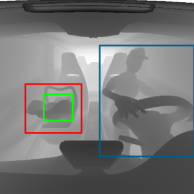} &
		\includegraphics[width = 0.15\textwidth]{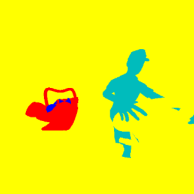} \\
	\end{tabular}
  \caption{Top : Real images. IR-image, depth image with 2D bounding box annotation, segmentation mask. \\ Bottom: Synthetic images. RGB scene, depth image with 2D bounding box annotation, segmentation mask.}
  {\label{fig:dataexamples}}  
\end{figure}

%brief comparison with other datasets
\begin{table*}[]
    \small
    \centering
    \begin{tabular}{|c||c|c|c|c|c|c|}
        \hline
        \textbf{Dataset}&\textbf{TICaM (Ours)}&\textbf{SVIRO\cite{cruz2020sviro}}&\textbf{AUC\cite{abouelnaga2018DDR}}&\textbf{Brain4Cars\cite{Brain4Cars}}&\textbf{HEH\cite{HEH}}&\textbf{Drive\&Act\cite{drive_and_act_2019_iccv}}\\
        \hline \hline
        \textbf{\makecell{Year}}&{2021}&{2020}&{2018}&\makecell{2016}&{2014}&{2019}\\
        \hline
        \textbf{\#Frames}&{$>$126K}&{25K}&{17K}&{2M}&{11K}&{\textbf{$>$9.6M}}\\
        \hline
        \textbf{\#Subjects}&{13}&{N/A}&{31}&{10}&{8}&{\textbf{15}}\\
        \hline
        \textbf{\#Views}&{1}&{1}&{1}&{2}&{2}&{\textbf{6}}\\
        \hline
        \textbf{Synthetic/Real}&{\textbf{Both}}&{Synthetic}&{Real}&{Real}&{Real}&{Real}\\
        \hline
        \textbf{Data}&{\textbf{Depth/RGB/IR}}&{\textbf{Depth/RGB/IR}}&{RGB}&{RGB}&{RGB/Depth}&{\textbf{Depth/RGB/IR}}\\
        \hline
        \textbf{Annotation}&\makecell{\textbf{2D+3D box,}\\\textbf{3D segmentation}\\ \textbf{mask, activity}}&\makecell{\textbf{classification labels,}\\\textbf{2D box and mask}\\ \textbf{2D keypoints}}&{activity}&{activity}&{activity}&{activity, 3D skeleton}\\
        \hline
        \textbf{\#Activity classes}&{20}&{N/A}&{10}&{5}&{3}&{\textbf{83}}\\
        \hline
        \textbf{Scenarios}&\makecell{\textbf{Driver}\\\textbf{Driver+Passenger}\\\textbf{Driver+Object}\\\textbf{Driver+Child seat}}&\makecell{\textbf{Driver}\\\textbf{Driver+Passenger}\\\textbf{Driver+Object}\\\textbf{Driver+Child seat}}&{Driver}&{Driver}&{Driver}&{Driver}\\
        \hline
        
    \end{tabular}
    \caption{Overview of existing in-car cabin monitoring datasets.}
    \label{tab:dataset_comparison}
\end{table*} 
An immediately apparent limitation of contemporary car cabin datasets is the lack of representation of some commonly occurring driving scenarios. For example, scenarios involving passengers, everyday objects, and child and infant in forward facing and rearward facing child seats respectively are missing in these datasets. We take care to record a wide range of driving scenarios that occur in everyday life so that our dataset can be used for critical automotive safety applications like airbag adjustment. Using the same airbag setting for a child as for an adult can be fatal for the child therefore, it is important to know the occupant class (person, child, infant, object or empty) of each car seat and to determine the child seat configuration(forward facing FF or rearward facing RF) \cite{graham1998reducing}. Our dataset contains also annotations of both driver and passenger activities. Activity recognition is not only crucial for innovative contact-less human machine interface but can moreover be fused with other modalities like driver gaze monitoring to obtain a robust estimation of the driver state, activity, awareness and distraction, which is crucial for hand-over maneuvers in conditional or highly automated driving \cite{Braunagel2017,Lex2015}.%While current systems rely mostly on a had gestures....recognition of hand g that predict user intentions. 

Another key feature missing from popular in-car cabin datasets is the multi-modality of them \cite{abouelnaga2018DDR,Brain4Cars}. We provide depth, RGB and infrared images with focus on 3D data annotations. These images have been captured inside a driving simulator using a Kinect Azure \cite{azure} fixed in the front near the rear-view mirror, providing a more practical viewpoint compared to other datasets \cite{abouelnaga2018DDR} and a mounting position that can be realistically replicated inside cars.

We use Time-of-Flight depth modality because it is associated with some unique benefits. Depth images preserve privacy as subjects cannot be identified, they are more robust to illumination and color variations, and allow natural background removal. Additionally, it is easier to generate realistic synthetic depth data for training machine learning systems than it is to generate RGB data, which is also something we provide as part of our dataset. For depth images we provide 2D and 3D bounding boxes, and class and instance masks. These annotations can be used as they are for training on infrared images and after some pre-processing on RGB images as well since the relative rotation and translation between depth and RGB cameras are known and provided. Lastly, we also provide annotations for activity recognition task making our dataset truly comprehensive in terms of input modalities and ground truth annotations.

Compared to real datasets, creation and annotation of synthetic datasets is less time and effort expensive. Recognizing this fact has led us to also create a synthetic front in-car cabin dataset of over 3.3K depth, RGB and infrared images with same annotations for detection and segmentation tasks. We believe this addition makes our dataset uniquely useful for training and testing car cabin monitoring systems leveraging domain adaptation methods.

To summarize, the main contributions of this paper are:
\begin{itemize}
    \item \textbf{TICaM}, a large-scale ToF dataset for in-car cabin monitoring consisting of a total of 6668 images annotated with 2D and 3D bounding boxes, segmentation masks, and over 123K RGB images with activity labels.
   
    \item  Additionally to real images, the \textbf{TICaM} contains 3306 synthetic images (Depth, RGB and IR imitation) with annotations for 2D and 3D object detection as well as instance segmentation, allowing for evaluation of domain adaptation methods.
    \item \textbf{TICaM} provides a new public benchmark for in-car cabin monitoring with high quality annotations, larger amounts of data and novel, more practical scenarios compared to existing datasets.
     \item The data in \textbf{TICaM} are recorded with a wide-angle RGB-D camera monitoring the entire front space of the cabin, capturing both driver and passenger seats.
    
\end{itemize}

%-------------------------------------------------------------------------
\section{Related Work}
Survey of existing car and driving datasets indicates that the most common application for such datasets is driver monitoring. Therefore, many datasets exist for tasks like driver distraction recognition \cite{abouelnaga2018DDR}, driver behavior recognition \cite{drive_and_act_2019_iccv}, driver gaze detection \cite{selim2020autopose,Ribeiro2019DriverGZ}, driver activity recognition \cite{HEH} and driver intention prediction \cite{Brain4Cars}. All these datasets provide color images of driver from front view for predicting driver activity or intention, with the exception of HEH and Dive\&Act which provide depth images as well. Brain4Cars \cite{Brain4Cars} dataset is used to classify driver intention into 5 categories for maneuver prediction. HEH \cite{HEH} captures additionally images of driver's hands for predicting driver activity. The AUC Distracted Driver dataset \cite{abouelnaga2018DDR} offers color images of drivers from side-view and can be used for detecting when the driver is distracted. All these datasets are limited in the number of image modalities they provide, with only one dataset providing depth images and none providing IR images. Moreover they are limited in the number of activity classes and image viewpoints they provide. Drive\&Act \cite{drive_and_act_2019_iccv} contributed a large-scale driver activity dataset with 6 views and multi-modal image data(RGB, Depth and IR) with 83 activity classes. However, their dataset provides activity and 3D skeleton annotations and is limited in terms of recorded scenarios. In contrast, TICaM provides multi-modal images and annotations for a wider range of driving scenarios with a single wide-angle front view that effectively captures the entire car cabin.

% Monitoring driver's gaze can give important cues about driver's attention and intention. For this we have datasets like
%  M Selim et al. “AutoPOSE: Large-scale Automotive Driver Head Pose and Gaze Dataset with Deep Head Orientation Baseline.” VISIGRAPP (2020): provides RGB, depth and IR from 2 camera views
 
%   R. F. Ribeiro and P. Costa, “Driver Gaze Zone Dataset With Depth Data”, IEEE International Conference on Automatic Face \& Gesture Recognition (FG 2019): gaze zone estimation for Drivers' inattention and distraction detection, rgb, depth and IR
  
 %Comparison with occupant classification datasets 
As mentioned in section \ref{Intro}, occupant classification is a safety-critical task for safe deployment of airbags. Many past works have addressed this problem but no dataset has been made publicly available \cite{OC1,OC2,OC3,OC4}. Nowruzi \etal~\cite{Nowruzi2019} released a dataset of thermal images for occupant classification, however, their dataset lacks images captured with child seats and children/infants in the scene. SVIRO \cite{cruz2020sviro} released a dataset of synthetic RGB, IR and depth images of interior of ten different vehicles for occupant classification into four classes: person (also including child and infant), child seat, infant seat and object. We borrow methods and materials from SVIRO to build the synthetic imageset of our dataset. To the best of our knowledge, no publicly released real dataset for occupant classification provides scenes with passengers, children and child seats. We refer readers to Table \ref{tab:dataset_comparison} for a summary of the differences between TICaM and other popular in-car cabin datasets.

%-------------------------------------------------------------------------

\section{TICaM}
We describe here in detail the dataset recording and annotation process for TICaM. Data capturing setup is described in section \ref{Data Capturing Setup}, the acquisition and rendering process of real and synthetic data is elaborated on in sections \ref{data acq} and \ref{synthetic rendering} respectively, the data format and the ground truth annotation format are described in sections \ref{data format} and \ref{anno format} respectively, and a summary of training and testing splits is provided in section \ref{data stats}.

\subsection{Data Capturing Setup}
\label{Data Capturing Setup}
\begin{figure}
    \centering
    %\vspace{-3cm}
    \includegraphics[width=0.45\textwidth]{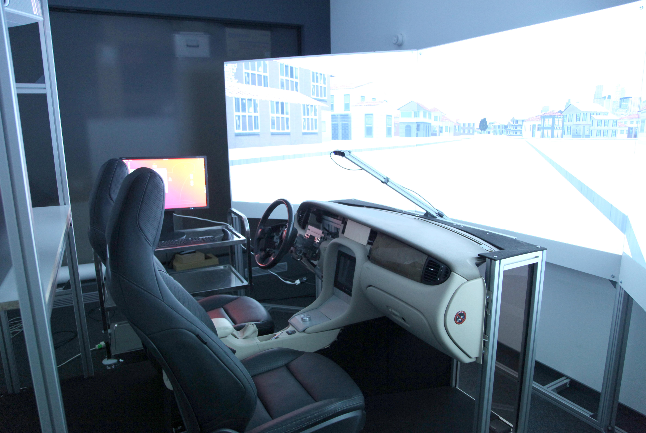}
    \caption{Our data capturing setup equipped with a wide-angle projection system, car front seats and a Kinect Azure camera in the front.}
    \label{fig:DCsetup}
\end{figure}

\begin{figure*}[h]
   	\centering
	\setlength{\tabcolsep}{0pt}
 	\begin{tabular}{ccccccccccc}
 		\includegraphics[height=2.7cm, width=2.7cm]{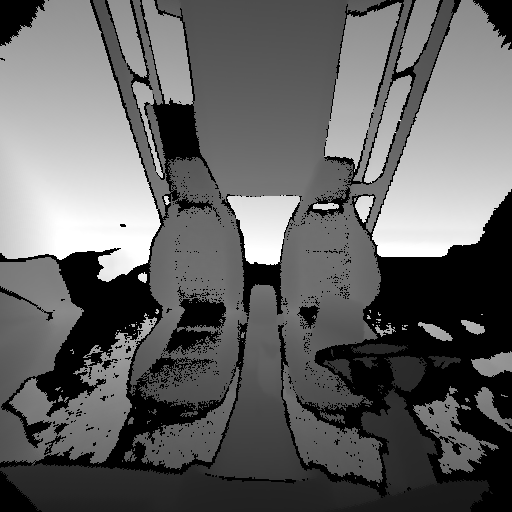}& {   } &
 		\includegraphics[height=2.7cm, width=2.7cm]{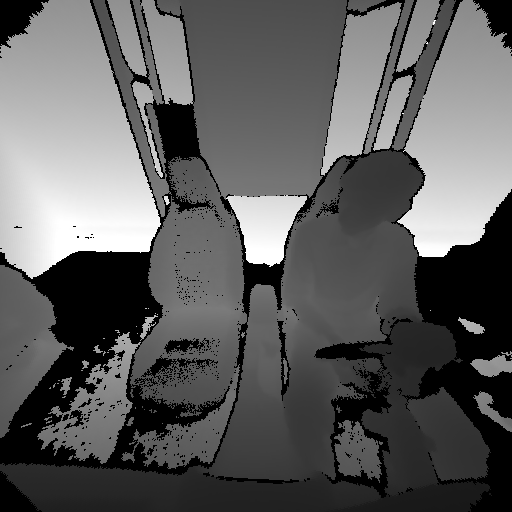}& {   } &
 		\includegraphics[height=2.7cm, width=2.7cm]{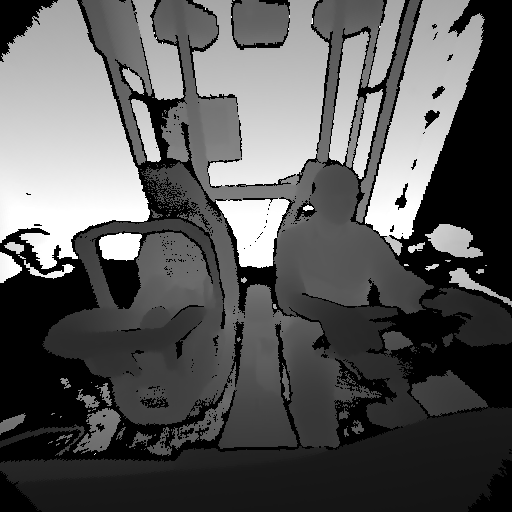}& {   } &
 		\includegraphics[height=2.7cm, width=2.7cm]{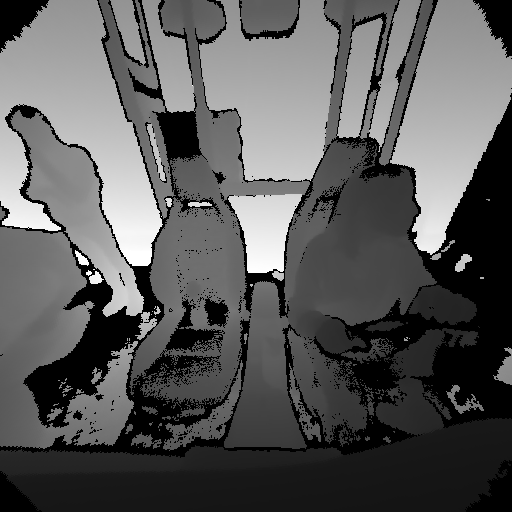}& {   } &
 		\includegraphics[height=2.7cm, width=2.7cm]{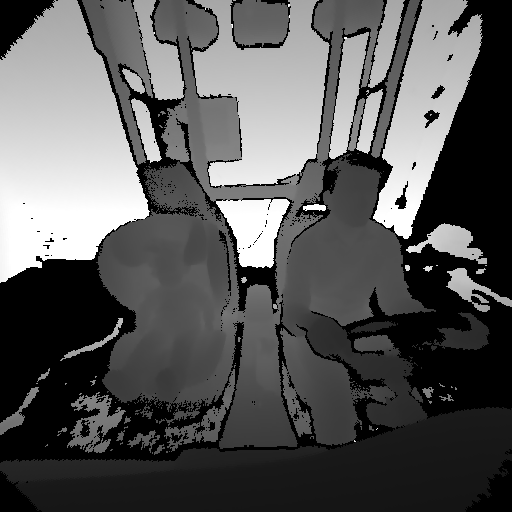}& {   } &
 		\includegraphics[height=2.7cm, width=2.7cm]{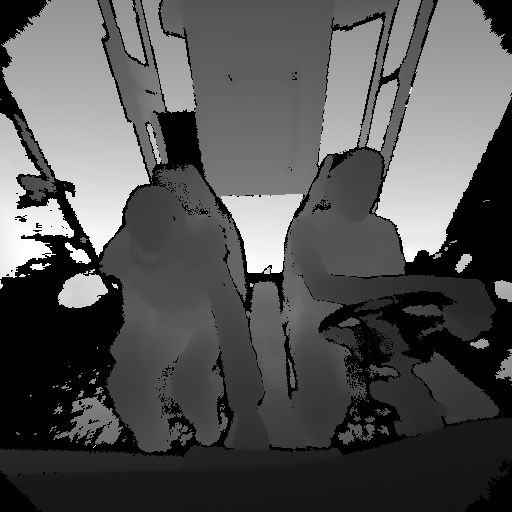}
       \\
 		\includegraphics[height=2.7cm, width=2.7cm]{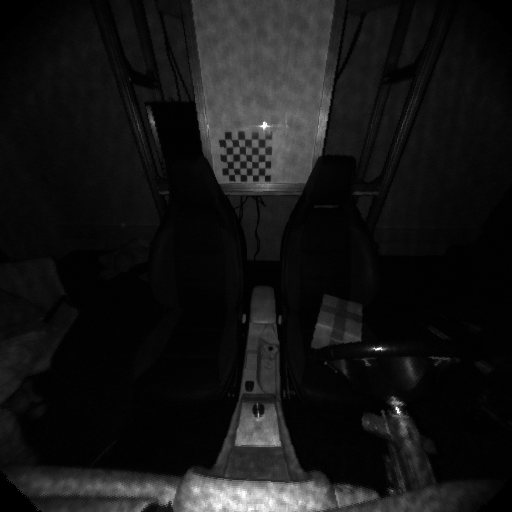}& {   } &
 		\includegraphics[height=2.7cm, width=2.7cm]{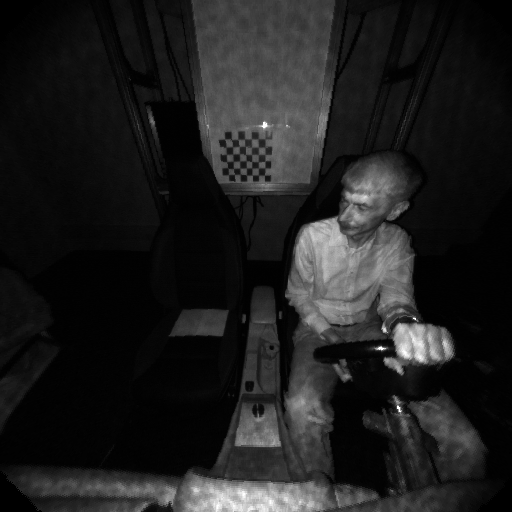}& {   } &
 		\includegraphics[height=2.7cm, width=2.7cm]{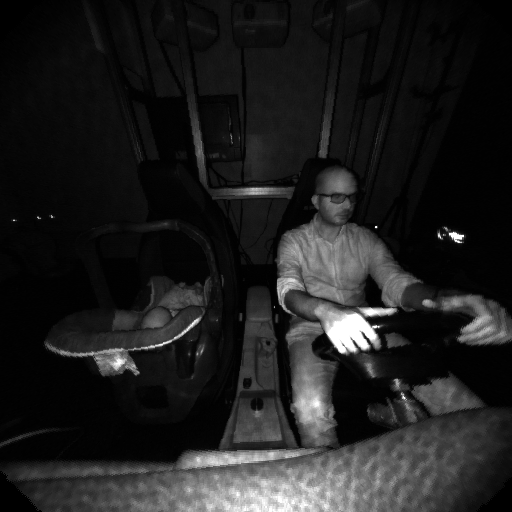}& {   } &
 		\includegraphics[height=2.7cm, width=2.7cm]{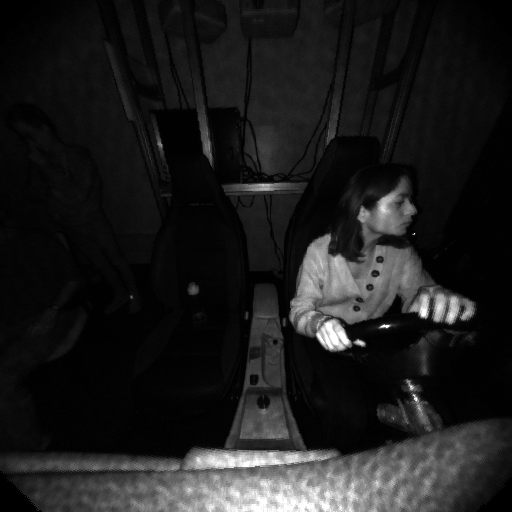}& {   } &
 		\includegraphics[height=2.7cm, width=2.7cm]{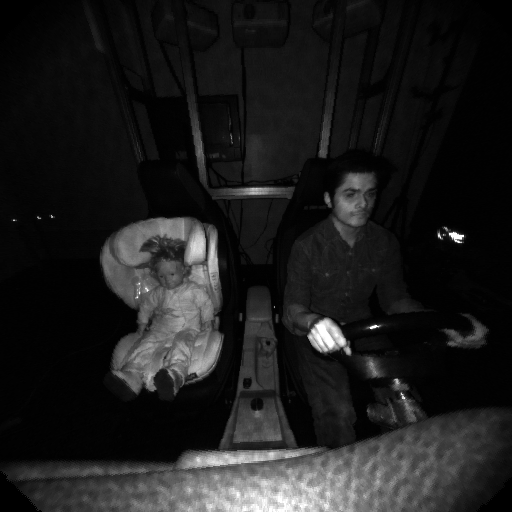}& {   } &
 		\includegraphics[height=2.7cm, width=2.7cm]{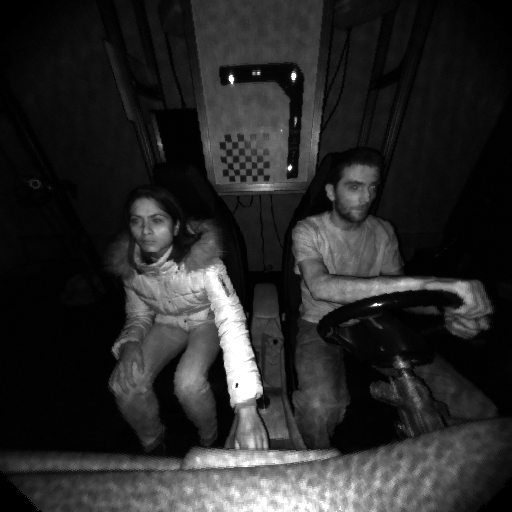}
 		\\
       \makecell{a) Object on \\ driver seat}& & \makecell{b) Only driver} & & \makecell{c) Driver and \\infant in RF} & & \makecell{d) Driver and \\ object} & & \makecell{e) Driver and \\ child in FF} & & \makecell{f) Driver and \\ Passenger}

       \\
 	\end{tabular}
  \caption{Depth and IR images of different driving scenarios provided in TICaM.}
  \label{scenarios} 
\end{figure*}

\begin{figure*}[h]
   	\centering
	\setlength{\tabcolsep}{0pt}
 	\begin{tabular}{ccccccccccc}
 		\includegraphics[height=2.7cm, width=2.7cm]{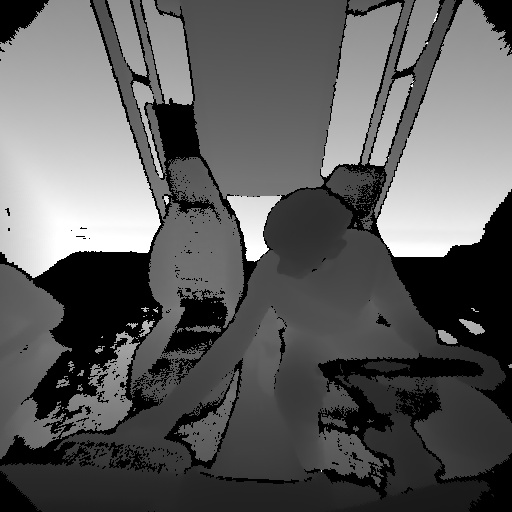}& {   } &
 		\includegraphics[height=2.7cm, width=2.7cm]{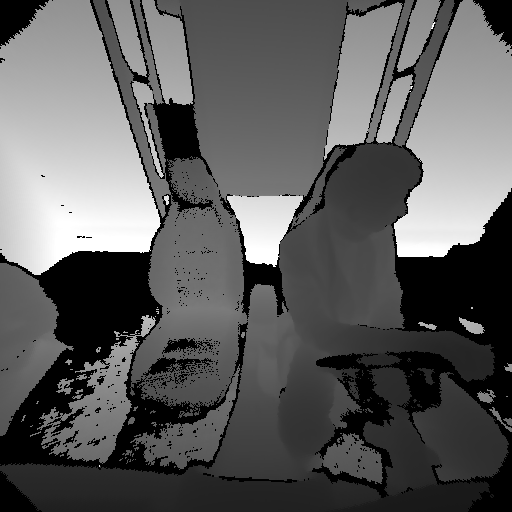}& {   } &
 		\includegraphics[height=2.7cm, width=2.7cm]{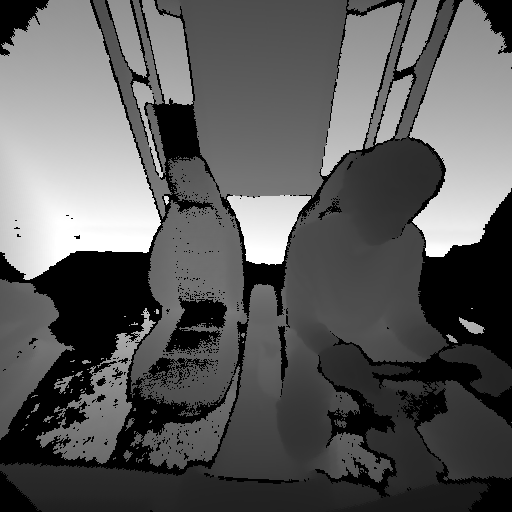}& {   } &
 		\includegraphics[height=2.7cm, width=2.7cm]{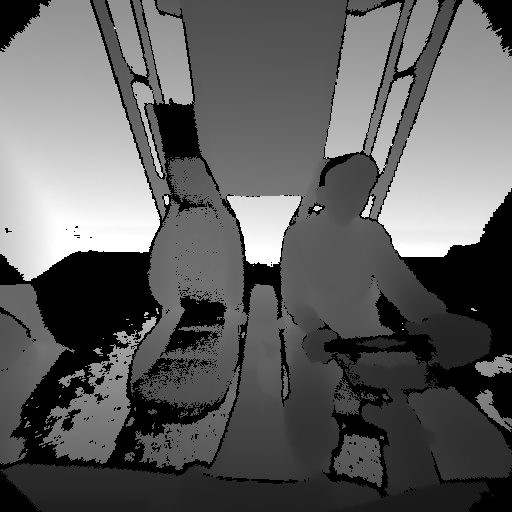}& {   } &
 		\includegraphics[height=2.7cm, width=2.7cm]{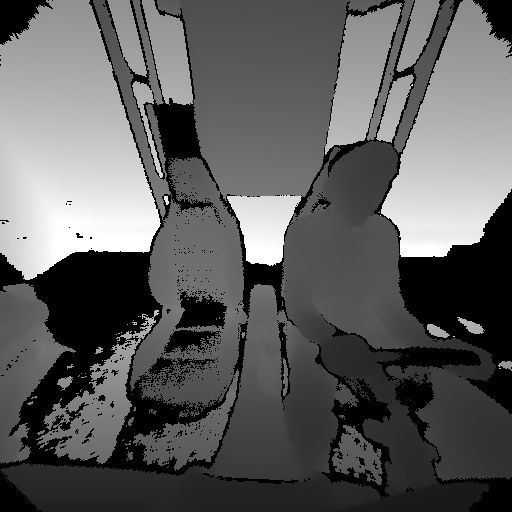}& {   } &
 		\includegraphics[height=2.7cm, width=2.7cm]{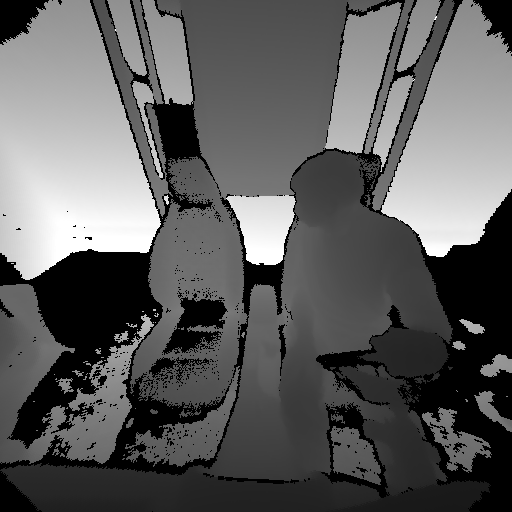}       \\
 		\includegraphics[height=2.7cm, width=2.7cm]{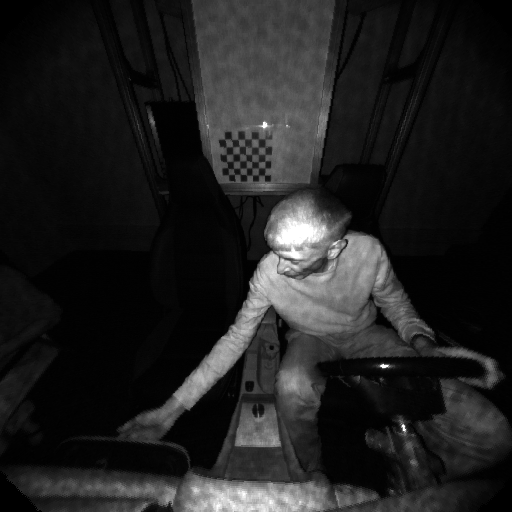}& {   } &
 		\includegraphics[height=2.7cm, width=2.7cm]{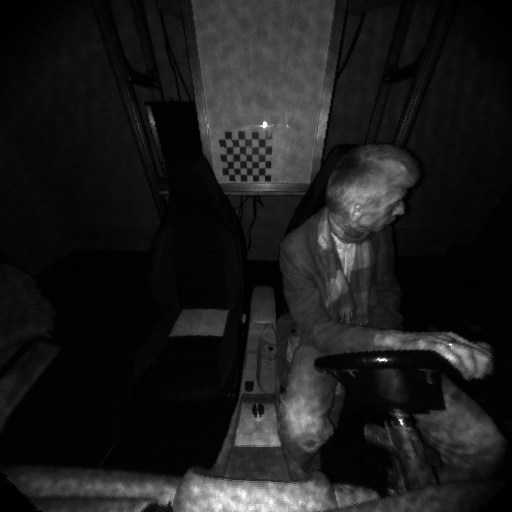}& {   } &
 		\includegraphics[height=2.7cm, width=2.7cm]{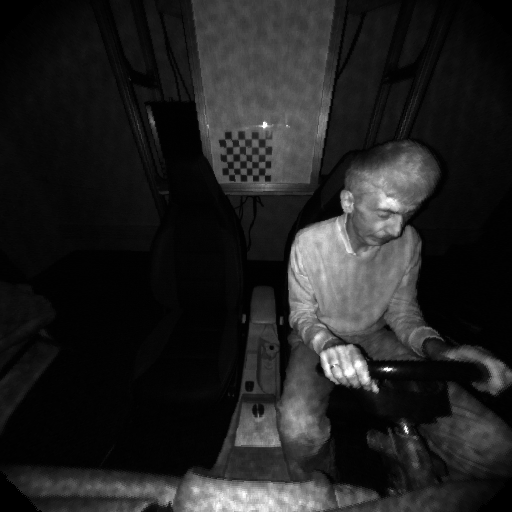}& {   } &
 		\includegraphics[height=2.7cm, width=2.7cm]{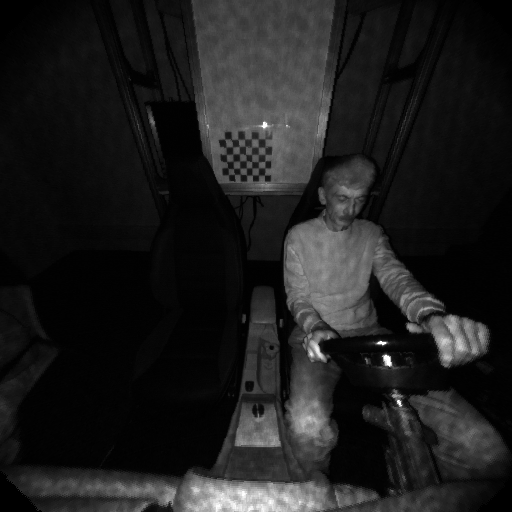}& {   } &
 		\includegraphics[height=2.7cm, width=2.7cm]{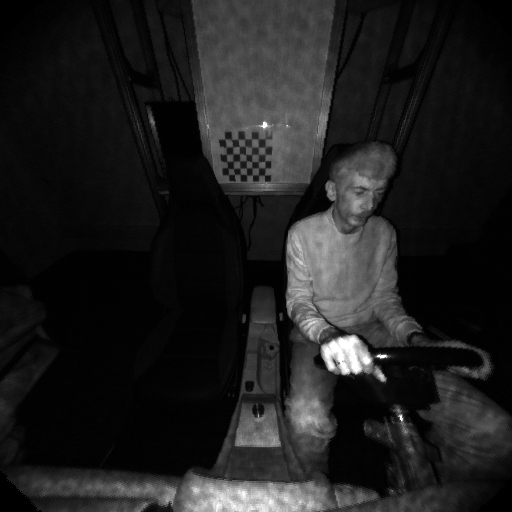}& {   } &
 		\includegraphics[height=2.7cm, width=2.7cm]{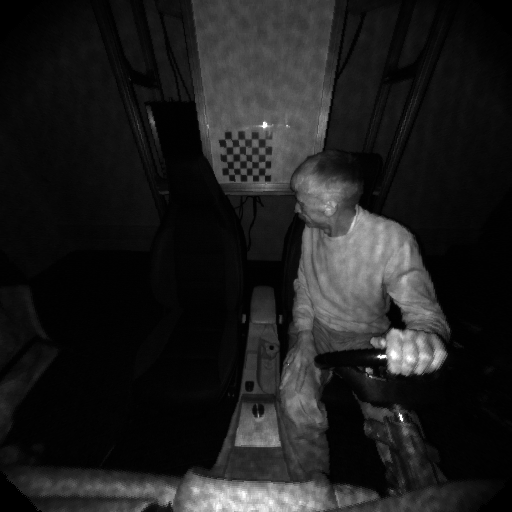}
       \\
       (a) && (b) && (c) && (d) && (e) && (f)

       \\
 	\end{tabular}
  \caption{We record every participant in several car seat positions and with varying accessories like jackets and hats. Figure 4(a) and 4(b) show a person in different clothing accessories. 4(c) and 4(d) show the difference in driver seat positions. Different driving actions are shown through Figure 4(a)-4(f).}
  \label{position} 
\end{figure*}

For data recording, we use an in-car cabin test platform developed by Feld et al.\cite{Feld2020}, shown in Figure \ref{fig:DCsetup}. It consists of a realistic in-car cabin mock-up, equipped with a wide-angle projection system for a realistic driving experience. A Kinect Azure camera with a wide field of view is mounted at the rear-view mirror position for 2D and 3D data recording. The camera is set to record at 30fps with $2\times2$ binning. The captured data consists of RGB, depth and IR amplitude images. To ensure a wide range of variability in the dataset, we adjust the seat positioning of the driver and passenger seats in the driving simulator via a CAN bus system for every recorded sequence.

\subsection{Data Acquisition}
\label{data acq}
We first define certain use cases or scenarios that we would like to be represented in our dataset (as shown in Figure \ref{scenarios}) and then record those scenarios with 13 participants, 4 female and 9 male. These scenarios include: 1) only driver in scene, 2) driver and passenger in scene, 3) driver and an object in scene, 4) driver and an empty Forward Facing Child Seat (FF) in scene, 5) driver and an empty Rearward Facing Infant Seat (RF) in scene, 6) driver and an occupied Forward Facing Child Seat (FF) in scene, 7) driver and an occupied Rearward Facing Infant Seat (RF) in scene, and 8) only an object in scene. We define a choreography each for the driver and the passenger which we share with the participants before recording of a driving sequence. For example, drivers performed actions like sitting, driving normally, looking left while turning wheel, turning right and so on. On the other hand, passengers performed actions like talking to the driver, grabbing something from the dashboard, etc. In total we have 20 actions for both driver and passenger. For each participant or pair of participants (in case of both driver and passenger), we record several sequences with varying positions of car seats so that any trained occupant classification and activity recognition model is robust to these differences. We also vary the appearance of the people through different clothing accessories like jackets and hats. Figure \ref{position} shows some examples of different seat positions, clothing accessories used and actions performed.

\begin{figure}[h]
  	\centering
	\setlength{\tabcolsep}{2pt}
	\begin{tabular}{cc}
		\includegraphics[height=2.7cm,width = 0.15\textwidth]{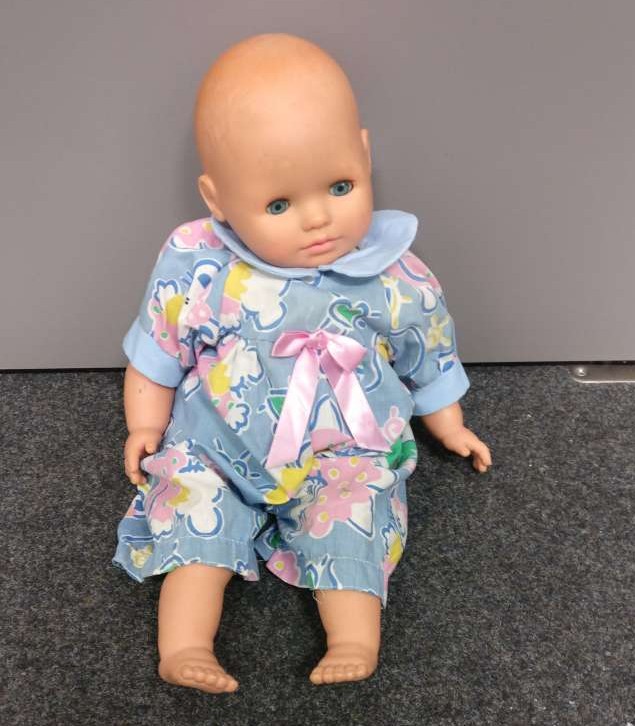} &
		\includegraphics[height=2.7cm,width = 0.15\textwidth]{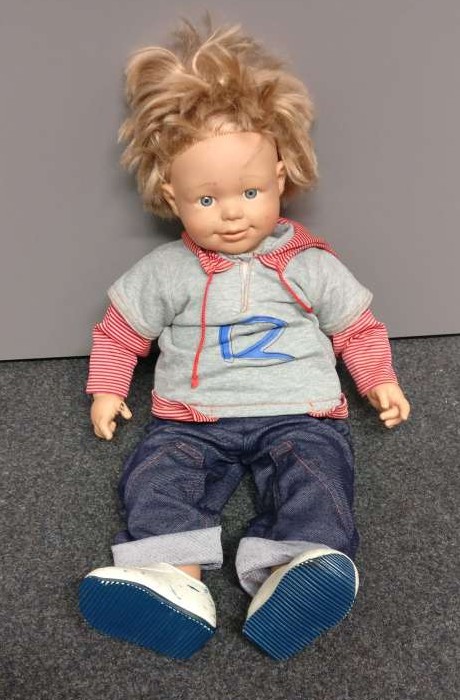} \\
		\includegraphics[height=2.7cm,width = 0.15\textwidth]{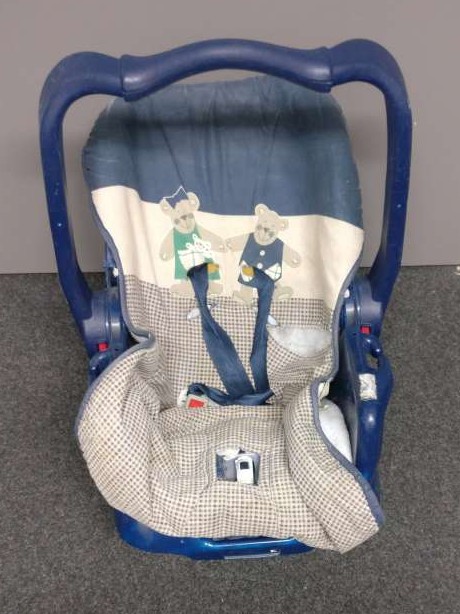} &
		\includegraphics[height=2.7cm,width = 0.15\textwidth]{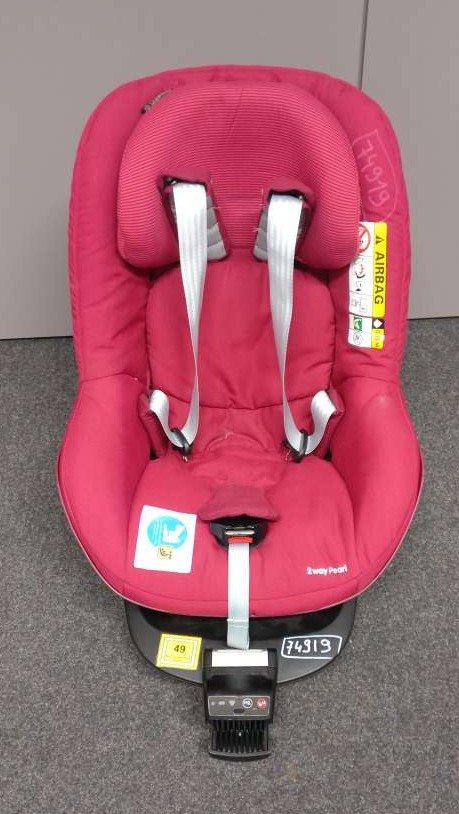} \\
	\end{tabular}
  \caption{Example human dolls and child seats used for recording scenarios with children and infants on the passenger seat.}
  {\label{dolls}}  
\end{figure}

For practical reasons we use human dolls as substitute for children and infants to record scenarios where one person is driving with an FF or RF seat securely placed on the passenger seat. Along with the dolls, we use 3 FF and 3 RF seats in different orientations like sun shade up/down or handle up/down. The human dolls and example RF and FF seats are shown in Figure \ref{dolls}.

\subsection{Synthetic Data Generation}
\label{synthetic rendering}
We refer to the materials and methods described in SVIRO \cite{cruz2020sviro} and vary the body poses of human models to adapt them for realistic driving poses to create our synthetic car in-car cabin dataset. The difference between SVIRO and TICaM being that SVIRO provides dataset for car rear-cabin while we provide dataset for car front-cabin. We render synthetic car cabin images using 3D computer graphic software Blender 2.81 \cite{blender} The 3D models of Mercedes A Class is from Hum3D \cite{hum3d}, the everyday objects were downloaded from Sketchfab \cite{sketchfab} and the human models were generated via MakeHuman \cite{makehuman}. In addition, High Dynamic Range Images (HDRI) \cite{hdri} was used to get different environmental backgrounds and illumination, and finally, in order to define the reflection properties and colors for the 3D objects, textures from Textures.com \cite{textures} were obtained for each object. 

\begin{figure}[h]
  	\centering
	\setlength{\tabcolsep}{2pt}
	\begin{tabular}{ccc}
		\includegraphics[width = 0.15\textwidth]{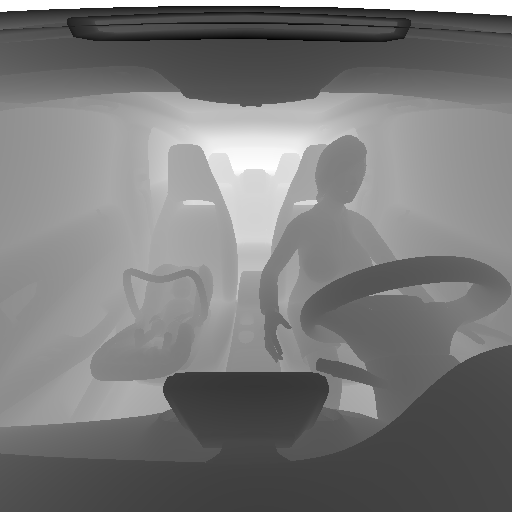} &
		\includegraphics[width = 0.15\textwidth]{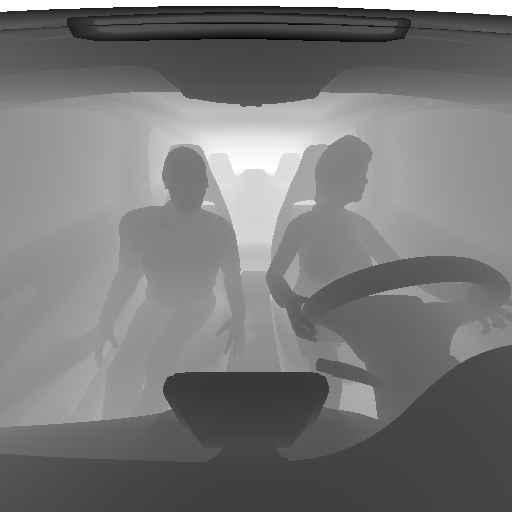} &
		\includegraphics[width = 0.15\textwidth]{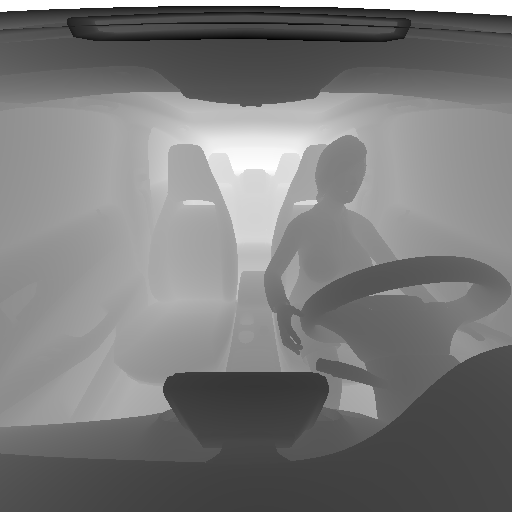} \\
		\includegraphics[width = 0.15\textwidth]{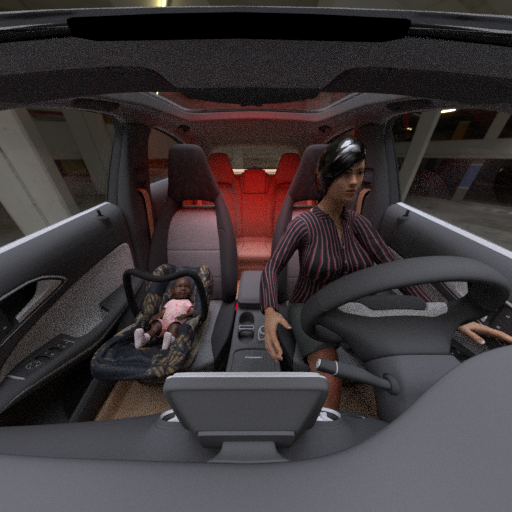} &
		\includegraphics[width = 0.15\textwidth]{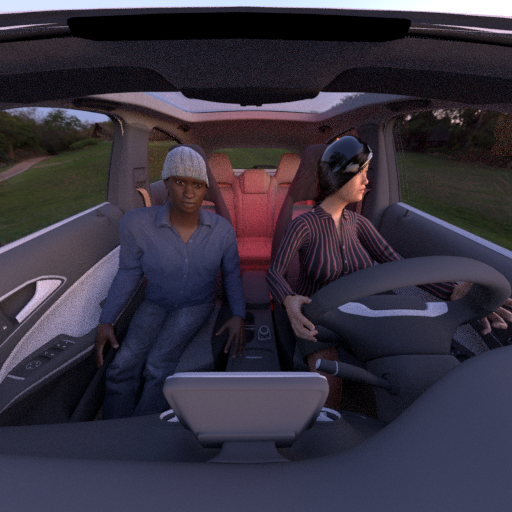} &
		\includegraphics[width = 0.15\textwidth]{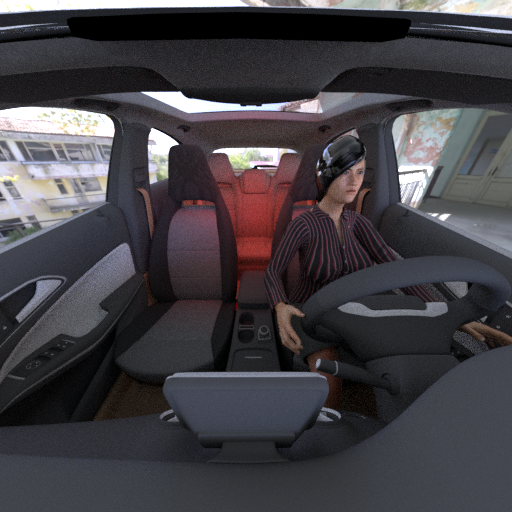} \\
	\end{tabular}
  \caption{Different scenarios in the synthetic dataset.}
  {\label{fig:syntheticscenarios}}  
\end{figure}

Since our goal was to generate simulated driving scenarios, we had the class occupying the
driving seat to always be an adult driver in a driving pose while for the passenger seat, we have randomly selected the occupant between the rest of the classes, i.e. child seats (empty and child-occupied), everyday objects, and adult passengers. We selected some of the same objects as the everyday objects in real driving scenarios like handbag, backpack, bottle,etc. but we group them under the same class 'object'. Similarly, children and infants are also grouped under class 'person'. We recreated the actions the participants were asked to perform during the recording of the real dataset. Since the driver poses are always restricted by the car elements they are interacting with, and in order to avoid intersection, we created some fixed poses for the hand positions that are possible in real driving scenarios, while allowing movement of other body parts up to some threshold. On the other hand, there were no such restrictions for the passenger poses, and thus we replicated the poses used in SVIRO by randomly selecting body poses within the physical constraints of the car seats. Figure \ref{fig:syntheticscenarios} shows a few images from the synthetic dataset showcasing the different scenarios present in this imageset.

\subsection{Data Format}
\label{data format}
Both the real and synthetic data are delivered in the same format for both image data and annotations with a few differences explained below. The camera data provided are in detail as follows:
\begin{itemize}
    \item \textbf{Depth Z-image}. The depth image is undistorted with a pixel resolution of $512\times512$ pixels and captures a $105^{\circ}\times105^{\circ}$ FOV. The depth values are normalized to $[1mm]$ resolution and clipped to a range of $[0,2550mm]$. Invalid depth values are coded as ‘0’. Images are stored in 16bit PNG-format.
    \item \textbf{IR Amplitude Image}. Undistorted IR images from the depth sensor are provided in the same format as the depth image above.
    \item \textbf{RGB Image}. Undistorted color images are saved in PNG-format in 24bit depth. While the synthetic RGB images have the same resolution and field of view as the corresponding depth images ($512\times512$), the real recorded RGB images have a higher resolution of $1280\times720$ pixels, but a lower field of view of $90^{\circ}\times59^{\circ}$ FOV. 
    \item \textbf{RGB Video}. The color video is provided as is recorded by the camera for sequences where a person is present in the scene. 
    \item \textbf{Camera intrinsics}. Along with the images, the camera intrinsics are also provided for both RGB and depth sensors in Kinect Azure as well as their relative rotation and translation. The camera intrinsics of the virtual camera in Blender are also provided.
    \item \textbf{Camera pose}. The different camera poses used for real recorded sequences are also provided as .yml files. 
\end{itemize}

\begin{table*}[h]
\caption{Overview of the object and activity classes present in TICaM.}
\label{classesOverview} \centering
\begin{tabular}{|c|c|c|c|}
  \hline
  \textbf{Data type} & \textbf{Annotation type} & \textbf{Class names} & \textbf{No. of Classes} \\
  \hline \hline
  \multirow{3}{*}{\textbf{Real}} &{Object classes} & {\makecell{Person,Backpack,WinterJacket,Box,WaterBottle,MobilePhone,\\
  Blanket,Accessory,Book,Laptop,LaptopBag,Infant,Handbag,FF,RF,Child}} & {15} \\
  \cline{2-4}
                             & {Activity Classes} &{\makecell{drive,look left while turning wheel,look right while turning wheel,\\touch screen,open glove compartment,touch head or face,\\lean forward,turn left,turn right,turn backwards while reversing,\\turn to front,adjust sun visor,turn backwards,talk,\\take something from dashboard,sitting normally,read paper or book,\\bending down,using phone,using laptop}} & {20} \\
  \hline \hline
  {\textbf{Synthetic}} & {Object Classes} & {\makecell{Person, FF, RF, Object}} & {4} \\
                        
  \hline
\end{tabular}
\end{table*} 

\subsection{Data Annotation}
\label{anno format}
The data annotation is done manually in principle, making use of the SALT 3D-annotation tool \cite{stumpf2021} that semi-automates the process through interpolation between frames and guided segmentation. We annotate every twentieth frame of the recorded sequences. For synthetic data, the ground truth is automatically generated by Blender. We provide the following annotations:
\begin{itemize}
    \item \textbf{2D bounding boxes}. For each real depth image , the 2D boxes are defined by the top-left and bottom-right corners of the box, its class label and a flag 'low remission' which is set to 1 for objects which are either black or very reflective or both, and therefore are barely visible in the depth image. Each 2D box in the synthetic dataset is represented by its class ID, the top-left and the bottom-right corners.
    \item \textbf{3D bounding boxes}. Each 3D box in the real dataset is represented by the coordinates (cx, cy, cz) of the box center, its dimensions (width, height, depth), its orientation along x-, y- and z- axes with respect to the world coordinate system, its class label and the 'low remission' flag. Whereas, each 3D box for the synthetic images is represented by the coordinates (cx, cy, cz) of the box center in camera coordinate system, its dimensions (width, height, depth) and its class ID.
    \item \textbf{Pixel segmentation masks}. For each real depth image two corresponding masks are generated: instance mask and class mask. The pixel intensities in these masks correspond to the class ID in the class mask and the instance ID for a certain class in the instance mask. For synthetic images we provide a single mask in the same format as in SVIRO \cite{cruz2020sviro}.
    \item \textbf{Activity annotations}. For all sequences with people in the driver or passenger seat, we provide a .csv file describing the activities performed throughout the sequence. Each .csv contains the activity ID, activity name, person ID, a label either 'Driver' or 'Passenger' to specify if the action is performed by the driver or he passenger, the starting frame number of the action, the ending frame and the duration of that action in frames.
\end{itemize}

\subsection{Data Statistics}
\label{data stats}
\begin{table}[h]
\caption{Number of frames in each of the real and synthetic imageset for training and testing for different tasks.}\label{traintest} \centering
\begin{tabular}{p{0.70cm}p{2.5cm}p{1.5cm}p{1cm}}
\hline
& {\makecell{\textbf{Activity} \\ \textbf{recognition}}} & \multicolumn{2}{c}{\makecell{\textbf{2D detection}\\\textbf{3D detection}\\\textbf{Segmentation}}} \\ 
\cline{2-4}
& \textbf{Real} & \textbf{Real} & \textbf{Synthetic} \\ 
\hline
\hline
\textbf{Train} & 86K  & 4.7K & 3.3K \\ 
\hline
\textbf{Test} & 37K  & 2.0K & -\\ 

\hline
\end{tabular}
\label{pnvsimage}
\end{table}

TICaM is a combined dataset of 6.7K real time-of-flight depth and IR images, video recordings equivalent to over 123K real RGB frames, and 3.3K synthetic depth, IR and RGB images. Table \ref{classesOverview} provides an overview of the object and the activity classes annotated in the dataset for detection, segmentation and activity recognition tasks. We use synthetic dataset only for training purpose in order to evaluate its domain adaptation benefit. We split the real dataset into training and testing sets such that a participant, RF, FF or an object instance belongs to either training or testing set. Table \ref{traintest} summarizes the number of images present in each set.

%\section{Benchmark evaluation}
%Here we have to discuss the role of the synthetic data and to clearly define testing and training data splits.  Do we have an idea regarding 3D box prediction ?
%-------------------------------------------------------------------------
\section{Conclusion}
We present TICaM, a comprehensive time-of-flight in-car cabin monitoring dataset consisting multi-modal images and multi-purpose annotations. We capture essential driving scenarios missing from contemporary driving datasets in both real and synthetic imagesets, and annotate them for 2D and 3D detection, instance segmentation and activity recognition. Our dataset can be used for training cabin monitoring systems to provide both safety-critical functionalities like safe deployment of airbags, driver behaviour monitoring as well as comfort functions. Furthermore, the similarity of the real and the synthetic imagesets renders TICaM suitable for the testing of domain adaption approaches.

\textbf{Acknowledgement.} This work was partially funded within the Electronic Components and Systems for European Leadership (ECSEL) Joint Undertaking in collaboration with the
European Union's H2020 Framework Program and Federal Ministry of Education and Research of the Federal Republic of Germany (BMBF), under grant agreement 16ESE0424 / GA826600 (VIZTA).

The authors would also like to thank Steve Dias Da Cruz from IEE S.A. for providing his support and the source code used in the creation of synthetic imageset.

{\small
\bibliographystyle{ieee_fullname}
\bibliography{egbib}
}

\end{document}